\PassOptionsToPackage{dvipsnames}{xcolor}
\documentclass[letterpaper, 10 pt, conference]{ieeeconf}  

\IEEEoverridecommandlockouts                              

\usepackage{times}
\usepackage[pdftex]{graphicx}
\usepackage{subfigure}
\usepackage{amsmath,amssymb,amsopn,amstext,amsfonts}
\usepackage{cancel}
\usepackage[space]{cite}
\usepackage{pdfsync}
\usepackage{balance}
\usepackage{color}
\usepackage{mathtools}
\usepackage{bm}
\usepackage{amsbsy}
\usepackage{diagbox}
\usepackage{float}
\usepackage{epstopdf}
\usepackage{pifont}
\usepackage{fixltx2e}
\usepackage{multirow}
\usepackage{url}
\usepackage{verbatim}
\usepackage[linkcolor=black,citecolor=black,urlcolor=black,colorlinks=true]{hyperref}
\usepackage{soul,xcolor}
\usepackage{algorithm,algorithmic}
\usepackage{subfiles}
\usepackage{stackengine}
\usepackage{pgfplots}
\usepackage[dvipsnames]{xcolor}
\usepackage{filecontents}

\graphicspath{{./img_compress/}}
\DeclareGraphicsExtensions{.png,.jpg,.eps,.pdf,.jpeg}

\newcommand{\etal}{\textit{et al.}}
\newcommand{\vect}{\mathbf}
\newcommand{\matr}[1]{\bm{#1}}

\DeclareMathOperator*{\argmin}{arg\,min}
\renewcommand\dddot[1]{\stackMath\stackengine{1pt}{#1}{.\mkern-1mu.\mkern-1mu.}{O}{c}{F}{T}{S}}

\title{\LARGE \bf
APACE: Agile and Perception-Aware Trajectory Generation for Quadrotor Flights
}

\author{Xinyi Chen, Yichen Zhang, Boyu Zhou and Shaojie Shen%
\thanks{B. Zhou is with the School of Artificial Intelligence, Sun Yat-Sen University, Zhuhai, China. X. Chen, Y. Zhang and S. Shen are with the Department of Electronic and Computer Engineering, Hong Kong University of Science and Technology, Hong Kong, China. \newline
\indent \tt\footnotesize $\{$xchencq, yzhangec, eeshaojie$\}$@connect.ust.hk \newline
\indent {\tt\footnotesize zhouby23@mail.sysu.edu.cn}
}%
}

\begin{document}
\maketitle
\thispagestyle{empty}
\pagestyle{empty}

\begin{abstract}
Various perception-aware planning approaches have attempted to enhance the state estimation accuracy during maneuvers, while the feature matchability among frames, a crucial factor influencing estimation accuracy, has often been overlooked.
In this paper, we present \textbf{APACE}, an \textbf{A}gile and \textbf{P}erception-\textbf{A}ware traje\textbf{C}tory g\textbf{E}neration framework for quadrotors aggressive flight, that takes into account feature matchability during trajectory planning.
We seek to generate a perception-aware trajectory that reduces the error of visual-based estimator while satisfying the constraints on smoothness, safety, agility and the quadrotor dynamics.
The perception objective is achieved by maximizing the number of covisible features while ensuring small enough parallax angles.
Additionally, we propose a differentiable and accurate visibility model that allows decomposition of the trajectory planning problem for efficient optimization resolution.
Through validations conducted in both a photorealistic simulator and real-world experiments, we demonstrate that the trajectories generated by our method significantly improve state estimation accuracy, with root mean square error (RMSE) reduced by up to an order of magnitude.
The source code will be released to benefit the community\footnote{\url{https://github.com/HKUST-Aerial-Robotics/APACE}}. 
\end{abstract}

\section{Introduction}
\label{sec:intro}

Visual onboard sensing has been popular on micro aerial vehicle platforms among perceptual modalities, thanks to recent advances in visual-inertial state estimation algorithms and camera's advantages of lightweight, small size, affordable price, and low power consumption.
For visual-based state estimation algorithms, such as VINS \cite{qin2018vins, qin2019general} and ORB-SLAM2 \cite{murORB2}, the accuracy of the estimator is significantly influenced by the captured image sequence.
For instance, a perception-agnostic trajectory risks facing texture-less regions (e.g. a white wall), where diminished availability of features deteriorates state estimation and potentially results in catastrophic collisions.
This motivates the development of \textit{perception-aware planning}, which takes into consideration the impact of perception when planning trajectories, for safe and reliable maneuvers.

Although numerous works on perception-aware planning have been proposed, there are still some limitations under aggressive flight scenarios, which may lead to disastrous state estimation failure.
Firstly, the aspect of feature matchability is often overlooked.
Some approaches \cite{Zhang2018, Costante2018} only focus on maximizing feature information gain, which implicitly assumes that a feature contributes to perception as long as it is visible.
However, only those matched and triangulated features can yield relative transformation information and improve state estimation.
Thus this assumption may not always hold, especially for high-speed maneuvers where the visual sensor captures rapidly changing images and the feature matchability among frames cannot be guaranteed.
Secondly, there is a lack of systematic, flexible and computationally efficient perception-aware planning frameworks for quadrotor aggressive flights that simultaneously address perception objectives to improve state estimation accuracy while ensuring constraints on the smoothness, safety, agility and the quadrotor dynamics.
One essential reason behind this gap is the absence of an accurate and differentiable visibility model allowing optimization decomposition with respect to position and yaw, resulting in challenging and inefficient trajectory optimization.
Moreover, existing approaches considering feature matchability are mostly tailored towards tracking clustered features and may be inflexible to handle environments with features distributed in a more general pattern.


\begin{figure}[t]
	\centering
  \includegraphics[width=\columnwidth]{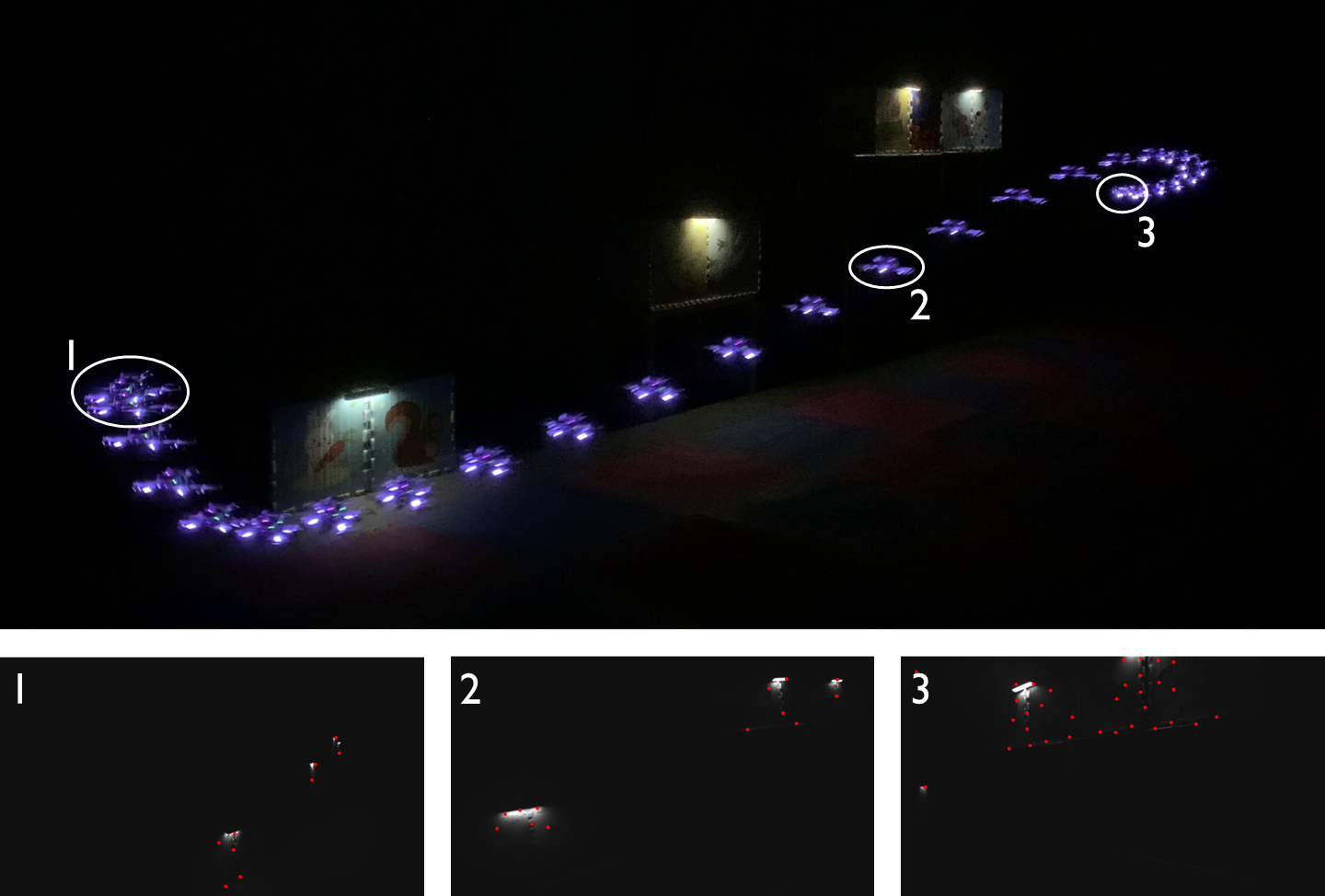}  
  \vspace{-0.6cm}
  \caption{\label{fig:realworld_large} A real-world aggressive flight experiment conducted in a challenging darkness environment. The LEDs on the quadrotor are only for photography purpose and disabled during the experiments. The bottom row shows snapshots of quadrotor's first-person view at the corresponding circled positions, where the red dots represent the matched features.}
  \vspace{-1.1cm}
\end{figure}

To address the above limitations, we present \textbf{APACE}, an \textbf{A}gile and \textbf{P}erception-\textbf{A}ware traje\textbf{C}tory g\textbf{E}neration framework for quadrotors in this paper.
Our motivation stems from the recognition that achieving accurate visual-based state estimation has two essential requirements:
(1) the parallax angles of covisible features observed by the quadrotor should be sufficiently small to ensure reliable feature association;
(2) the quadrotor should be able to observe adequate covisible features between consecutive frames for feature triangulation.
Hence we seek to generate trajectories that reduce the estimation error by ensuring small enough parallax angles while maximizing the number of covisible features. 
We observe that the parallax angle is independent of the yaw direction, which allows decomposing the optimization problem into two subproblems utilizing the differential flatness property of quadrotors.
We also propose a novel accurate and differentiable visibility model, which enables the objective of covisibility to be optimized in two steps.
This approach allows efficient resolution of the optimization problem, resulting in high-quality perception-aware trajectories for aggressive quadrotor flight.

Compared to previous works, our method is able to generate agile and perception-aware trajectory efficiently that achieves higher visual-based state estimation accuracy in diverse scenarios.
We benchmark and validate our method in a photorealistic simulator and conduct real-world experiments.
The result demonstrates that the trajectories generated by our method achieve higher state estimation accuracy and reliability during more aggressive maneuvers in challenging low-texture environments, reducing the root mean square error (RMSE) by up to an order of magnitude.
In summary, the contributions of this work are:
\begin{enumerate}
  \item A systematic, flexible and efficient perception-aware trajectory generation framework for quadrotor agile flight, which takes feature matchability into account.
  \item An accurate and differentiable visibility model to evaluate the number of covisible features, which allows decomposition of the trajectory planning problem.
  \item Method validation in a photorealistic simulator and real-world experiments. The source code will be released to benefit the community.
\end{enumerate}

\section{Related Work}
\label{sec:related}


Among recent perception-aware planning studies, information gain evaluated using various metrics has been used to bridge perception and planning.
Fisher information is utilized in \cite{Zhang2018, Kim2021, Costante2018} as the metric for perception quality, which quantifies the information that the observations carry about the desired state.
Bartolomei \etal \cite{Bartolomei2020} utilize semantic segmentation to avoid texture-less areas which may cause estimation failures.
Sabdat \etal \cite{Sadat2014} suggest a perception quality metric consisting of mesh triangle density and visible surfaces viewing angle and integrates this metric into Rapidly-Exploring Random Tree \cite{karaman2011}.
These works commonly overlook feature matchability, assuming any visible features contribute to perception.
However, this assumption may not always hold, especially in agile flight when the feature matchability between frames cannot be guaranteed.

Alternatively, feature matchability has been taken into consideration.
Sheckells \etal \cite{Sheckells2016a} optimize an execution trajectory minimizing the image feature reprojection error along the trajectory, given a goal image to align with.
Falanga \etal \cite{Falanga2018} introduce a perception-aware model predictive control (MPC) framework, which maximizes the visibility of points of interest and minimizes their velocities in the image frame.
Spasojevic \etal \cite{Spasojevic2020} propose a trajectory parameterization algorithm that minimizes the execution time while preserving a given set of features within the field of view.
Penin \etal \cite{Penin2017} solves a minimum-time optimization problem with a hard-constraint on keeping a certain set of features visible.
In \cite{Penin2018}, obstacle avoidance in the environment and vision occlusion in the image space are further considered.
However, the aforementioned works are commonly tailored for tracking clustered points of interest during the trajectory execution.
As a result, their algorithms are not flexible enough to deploy in scenarios with features distributed in more general patterns when one needs to maintain a reduced set of features within the field of view to hit desired agility.
Therefore, they may be unsuitable for our task where agent agility is preferred over target tracking.

Approaches that include feature matchability as a soft-constraint in the planning problem are also proposed.
Greeff \etal \cite{Greeff2020} propose a probabilistic model of feature matchability to be incorporated into an MPC chance constraint, ensuring the number of visible features is over a lower bound.
However, their method is designed for a teach-and-repeat framework, which differs from our task.
Murali \etal \cite{Murali2019} propose an optimization algorithm for yaw trajectory with a predetermined position trajectory that ensures both feature covisibility and motion agility.
However, perception quality can also be greatly influenced by the position trajectory, which should be considered in the optimization problem.

Various visibility functions have been proposed to model whether a feature lies in the camera fields of view.
An axially symmetric conical model is deployed in \cite{Zhang2019, Spasojevic2020, Penin2017}, which oversimplifies the view frustum and may not be capable of handling situations when horizontal and vertical fields of view are different.
In \cite{Murali2019}, the fields of view is modeled as the intersection of five half spaces, each of which corresponds to a facet of the view frustum.
However, its high complexity leads to slow optimization convergence.

\section{System Overview}
\label{sec:problem}

\begin{figure}[t]
	\begin{center}          
    {\includegraphics[width=\columnwidth]{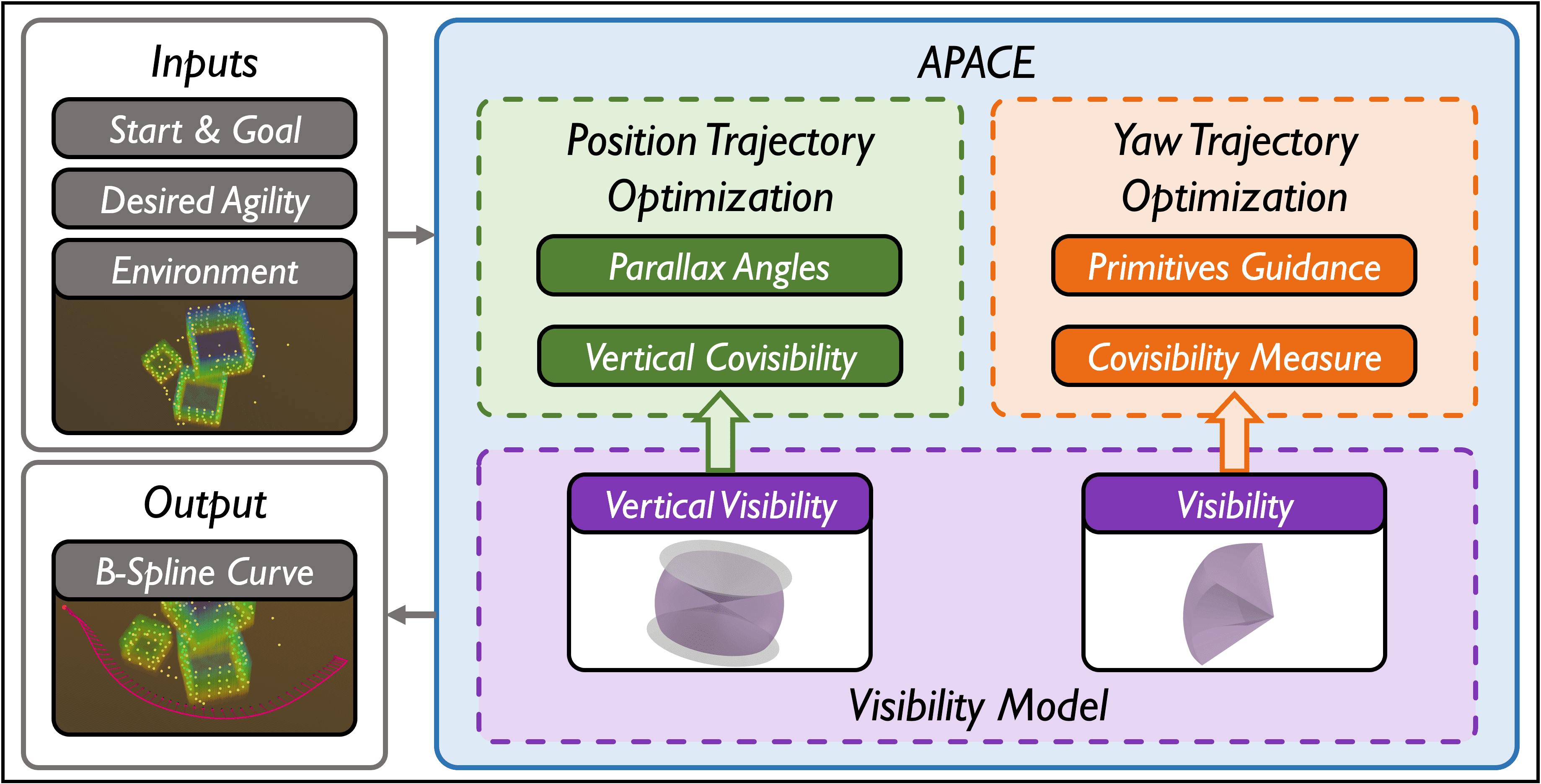}}       
  \end{center}
  \vspace{-0.4cm}
   \caption{\label{fig:overview} An overview of the proposed trajectory generation method.}
   \vspace{-0.8cm}
\end{figure}

Given start and goal positions and desired agility in terms of top velocity and environmental information including feature points and obstacles, we aim to generate a perception-aware trajectory that reduces the error of visual-based estimator while satisfying the constraints on smoothness, safety, agility and the quadrotor dynamics.
The perception objective is achieved by maximizing the number of covisible features and ensuring small enough parallax angles.
The position and yaw trajectories are represented in the form of uniform B-spline curves (Sec.\ref{subsec:bspline}), where the control points of these curves serve as the optimization variables.
Leveraging the independency of parallax angle from yaw direction (Sec.\ref{subsec:parallax}), the quadrotor's property of differential flatness (Sec.\ref{subsec:diff_flatness}) and the decomposable visibility model proposed in Sec.\ref{subsec:model}, we adopt a two-stage trajectory generation strategy, which includes position and yaw trajectory generation detailed in Sec.\ref{subsec:pos_traj} and Sec.\ref{subsec:yaw_traj} respectively.
The system overview is depicted in Fig.\ref{fig:overview}.

\section{Preliminaries}
\label{sec:prelim}
\subsection{Uniform B-spline Curves}
\label{subsec:bspline}
A B-spline curve $\mathcal{C}(t)\in\mathbb{R}^3$ is a piecewise polynomial curve which can be uniquely defined by its \textbf{degree} $p$, a set of $n+1$ \textbf{control points} $\{\vect{Q}_0, \vect{Q}_1, \cdots, \vect{Q}_n\}$ and a \textbf{knot vector} $T=(t_0, t_1, \cdots, t_m)$, where $\vect{Q}_i\in\mathbb{R}^3, t_j < t_{j+1}$ and $m = n + p + 1$.
A B-spline curve $\mathcal{C}(t)$ can be parameterized using time $t$, the domain of which is $[t_p, t_{m-p}]$. 
For a knot $t_j\in T, j = p, p+1, \dots, m-p$, the point $\mathcal{C}(t_j)$ on the curve corresponds to $t_j$ is referred as a \textbf{knot point}, giving the set of knot points $\{\vect{P}_p, \vect{P}_{p+1}, \cdots, \vect{P}_{m-p}\}$.
For a general time instance $t\in[t_k, t_{k+1}) \subset [t_p, t_{m-p}]$, the value $\mathcal{C}(t)$ can be evaluated using a matrix representation \cite{qin2000general}
\begin{equation}
\begin{aligned}
  \mathcal{C}(t) &= \matr{U}(t) \matr{M}_{p+1}(k) \matr{V}(k)^T, \\
  \matr{U}(t) &= \begin{bmatrix} 1 & u(t) & u^2(t) & \cdots & u^p(t) \end{bmatrix}, \\
  \matr{V}(k) &= \begin{bmatrix} \vect{Q}_{k-p} & \vect{Q}_{k-p+1} & \cdots & \vect{Q}_{k} \end{bmatrix},
\end{aligned}
\end{equation}
where $u(t) = (t - t_k)/(t_{k+1} - t_k)$ and $\matr{M}_{p+1}(k)$ is the $k$-th B-spline basis matrix of degree $p$.
For a \textbf{uniform} B-spline curve, the \textbf{knot spans} between consecutive knots are identical, i.e. $t_{j+1} - t_j = \Delta t$ and the basis matrix $\matr{M}_{p+1}(k)$ is a constant matrix determined by the degree $p$.

\subsection{Parallax Angle}
\label{subsec:parallax}
Parallax angle is defined as the angle between the rays from a feature $\vect{f}$ to two camera positions $\vect{c}$ and $\vect{c}^\prime$ at different time instances, describing the apparent displacement or difference of the feature as a result of the sensor positional offset.
The parallax angle is only related to the position of the sensor, and independent of its yaw direction.
A large parallax angle may lead to feature matching failure \cite{song2021view} and deteriorate state estimation accuracy, which should be avoided during the maneuvers.

\subsection{Differential Flatness}
\label{subsec:diff_flatness}
Mellinger \etal \cite{MelKum1105} show the differential flatness of quadrotor dynamics, which means the states and the inputs of the quadrotor can be expressed as algebriac functions of the flat outputs and their derivatives.
The flat outputs are selected as $\sigma = [x,y,z,\psi]$, where $\vect{p}_i = (x,y,z)$ is the position coordinate of the quadrotor and $\psi$ is its yaw rotation.
Differential flatness is often exploited to independently plan trajectories for the flat outputs, enabling the decomposition of the trajectory planning problem into position and yaw optimization.


\section{Methodology}
\label{sec:algorithm}  

\subsection{Visibility Model}
\label{subsec:model}

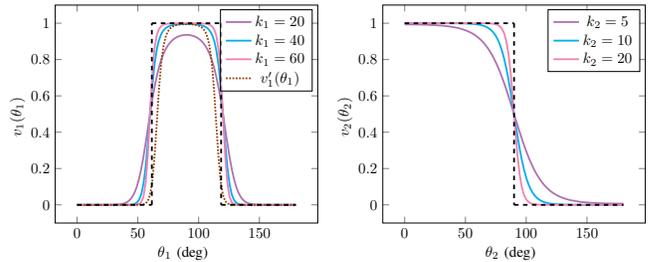
\begin{figure}
  \centering
  \resizebox{0.49\columnwidth}{!}{%
    \begin{tikzpicture}[declare function = {
      sigma1(\x)=1/(1 + exp(-20*(sin(\x) - 0.866)));
      sigma2(\x)=1/(1 + exp(-40*(sin(\x) - 0.866)));
      sigma3(\x)=1/(1 + exp(-60*(sin(\x) - 0.866)));
      sigma4(\x)=1/(1 + exp(-60*(sin(\x) - 0.9135)));}]

      \begin{axis}[
        scale=0.9, 
        legend style={fill=white, fill opacity=0.6, draw opacity=1, text opacity=1}]
      
        \addplot [domain=0:180, samples=100, color=Orchid, very thick] (x,{sigma1(x)});
        \addlegendentry{$k_1=20$}

        \addplot [domain=0:180, samples=100, color=Cyan, very thick] (x,{sigma2(x)});
        \addlegendentry{$k_1=40$}

        \addplot [domain=0:180, samples=100, color=CarnationPink, very thick] (x,{sigma3(x)});
        \addlegendentry{$k_1=60$}

        \addplot [domain=0:180, samples=100, color=RawSienna, very thick, densely dotted] (x,{sigma4(x)});
        \addlegendentry{$v_1^\prime(\theta_1)$}

        \draw[dashed, very thick] (axis cs:0,0)--(axis cs:61.5,0);
        \draw[dashed, very thick] (axis cs:61.5,0)--(axis cs:61.5,1);
        \draw[dashed, very thick] (axis cs:61.5,1)--(axis cs:118.5,1);
        \draw[dashed, very thick] (axis cs:118.5,1)--(axis cs:118.5,0);
        \draw[dashed, very thick] (axis cs:118.5,0)--(axis cs:180,0);
      \end{axis}

      \node at (3cm,-0.7cm) {$\theta_1$ (deg)};
      \node at (-0.8cm,2.5cm) [rotate=90, anchor=base] {$v_1(\theta_1)$};
    \end{tikzpicture}
  }
  \resizebox{0.49\columnwidth}{!}{%
    \begin{tikzpicture}[declare function = {
      sigma1(\x)=1/(1 + exp(-5*(cos(\x))));
      sigma2(\x)=1/(1 + exp(-10*(cos(\x))));
      sigma3(\x)=1/(1 + exp(-20*(cos(\x))));}]

      \begin{axis}[
        scale=0.9, 
        legend style={fill=white, fill opacity=0.6, draw opacity=1, text opacity=1}]
      
        \addplot [domain=0:180, samples=100, color=Orchid, very thick] (x,{sigma1(x)});
        \addlegendentry{$k_2=5$}

        \addplot [domain=0:180, samples=100, color=Cyan, very thick] (x,{sigma2(x)});
        \addlegendentry{$k_2=10$}

        \addplot [domain=0:180, samples=100, color=CarnationPink, very thick] (x,{sigma3(x)});
        \addlegendentry{$k_2=20$}

        \draw[dashed, very thick] (axis cs:0,1)--(axis cs:90,1);
        \draw[dashed, very thick] (axis cs:90,1)--(axis cs:90,0);
        \draw[dashed, very thick] (axis cs:90,0)--(axis cs:180,0);
      \end{axis}

      \node at (3cm,-0.7cm) {$\theta_2$ (deg)};
      \node at (-0.8cm,2.5cm) [rotate=90, anchor=base] {$v_2(\theta_2)$};
    \end{tikzpicture}
  }
  \vspace{-0.8cm}
  \caption{Visualization of the sigmoid functions (\ref{eq:v1}) and (\ref{eq:v2}) with $\alpha_v = 60^{\circ}$ and different $k$ values where the black dashed lines indicate the corresponding step functions to be approximated. The brown dotted curve in the left plot illustrates the constant weight $v_1^\prime(\theta_1)$ used in (\ref{eq:f_para}).}
  \vspace{-0.8cm}
  \label{fig:sigmoid}
\end{figure}

The field of view (FoV) of a camera is commonly expressed as an angle of view pair $(\alpha_h, \alpha_v)$, where $\alpha_h, \alpha_v$ are the horizontal and vertical FoV angles.
For a quadrotor state $\vect{s}_i=(\vect{p}_i, \psi_i), \vect{p}_i = (x_i,y_i,z_i)$ and a feature $\vect{f}_j$ in world frame, the exact visibility function takes value $1$ if feature $\vect{f}_j$ is within camera FoV at state $\vect{s}_i$ and $0$ otherwise.
We present a differentiable visibility model that approximates this exact visibility function more accurately and allows optimization decomposition of the covisibility objective for position and yaw trajectories.

Define $\vect{n}_1$ and $\vect{n}_y$ as the unit thrust direction of the quadrotor and the unit yaw direction in world frame respectively. Then
\begin{equation} \label{eq:n1_ny}
  \vect{n}_1 = \frac{\vect{a} - \vect{g}}{|\vect{a} - \vect{g}|}, \; \vect{n}_y = (\cos\psi, \sin\psi, 0)
\end{equation}
where $\vect{a}, \vect{g}$ and $\psi$ are the acceleration, gravity and the yaw angle.
For simplicity, we assume the quadrotor is equipped with a forward-looking camera and the optical axis of the camera $\vect{n}_2$ is orthogonal to the thrust direction $\vect{n}_1$. 
The model can be generalized to different camera configuration with minor adjustments.
Then $\vect{n}_2$ can be calculated as
$$\vect{n}_3 = \frac{\vect{n}_1 \times \vect{n}_y}{|\vect{n}_1 \times \vect{n}_y|}, \; \vect{n}_2 = \vect{n}_3 \times \vect{n}_1.$$

\begin{figure}[t]
  \centering
	\subfigbottomskip=-2pt
  \subfigure{\includegraphics[width=0.475\columnwidth]{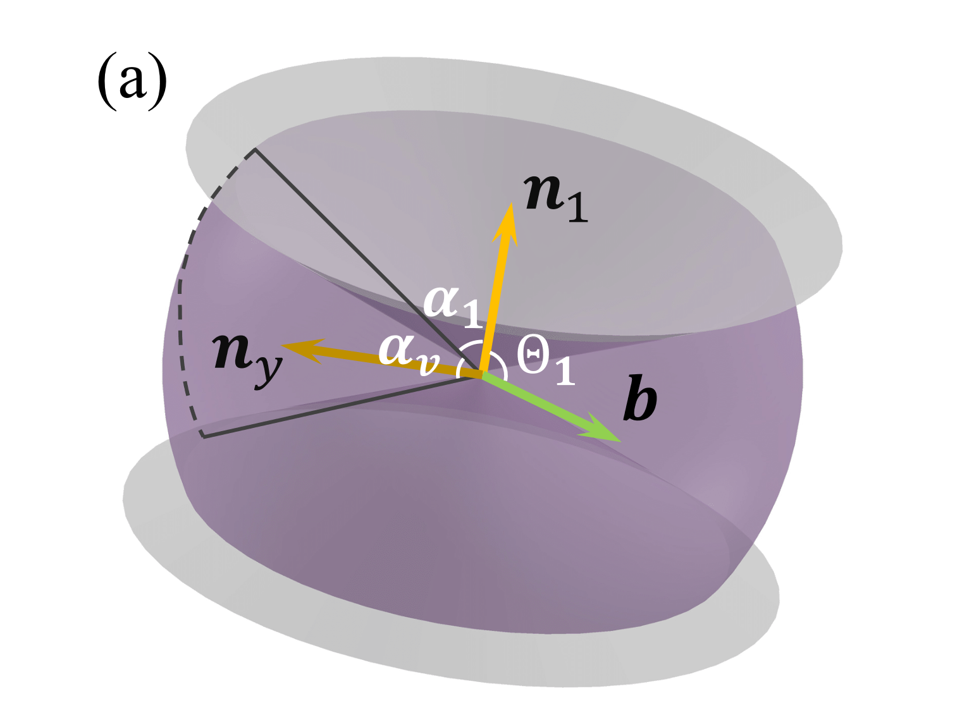} \label{fig:visib_model_1}}
  \subfigure{\includegraphics[width=0.475\columnwidth]{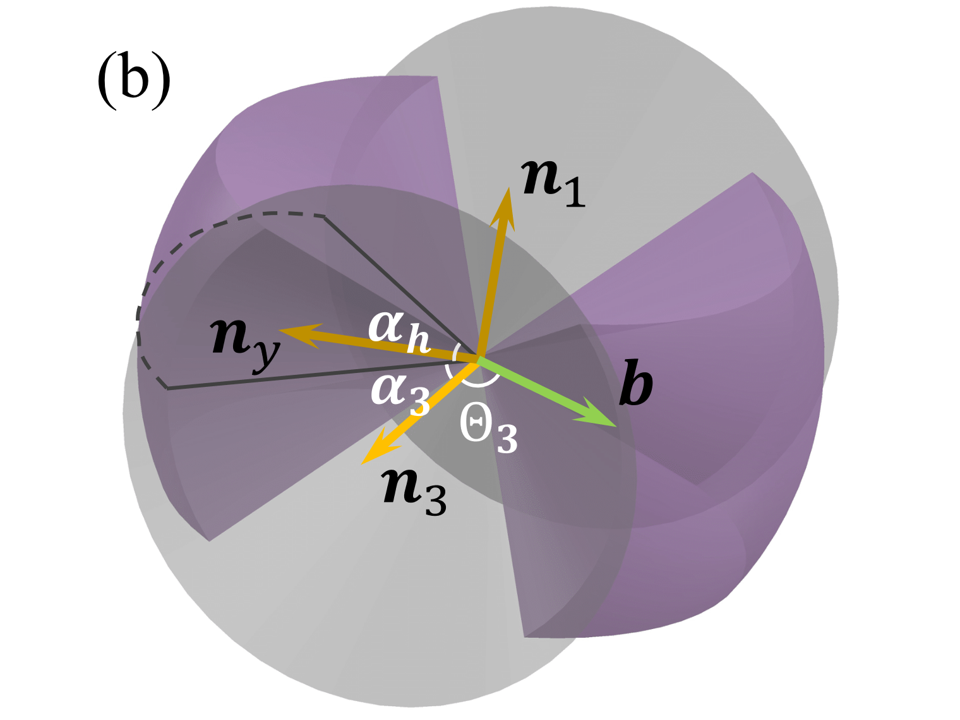} \label{fig:visib_model_2}}
  \subfigure{\includegraphics[width=0.475\columnwidth]{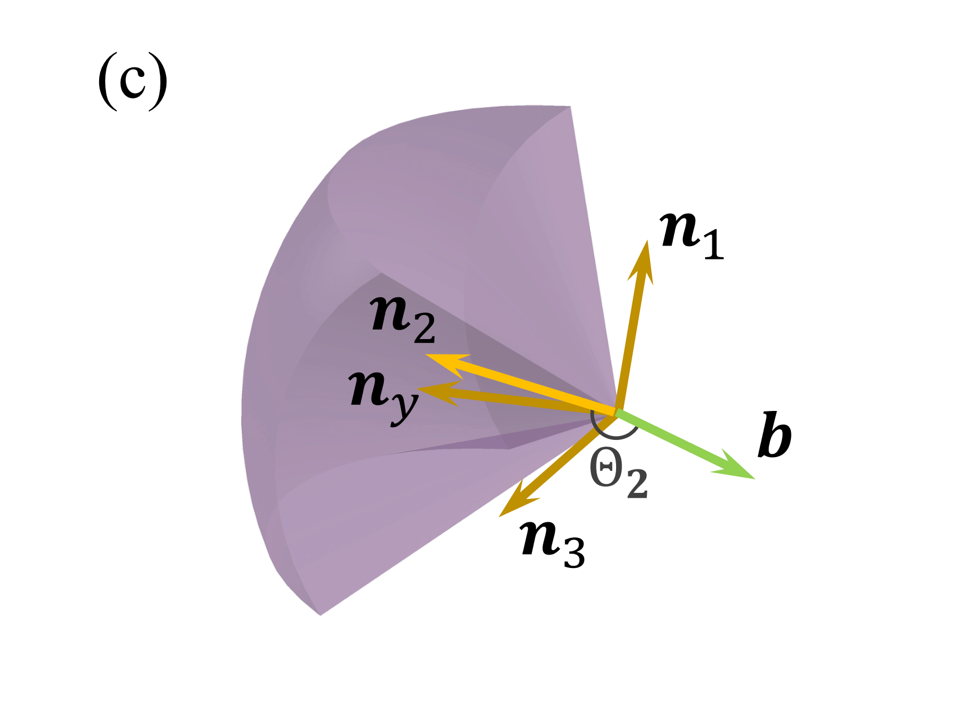} \label{fig:visib_model_3}}
  \subfigure{\includegraphics[width=0.475\columnwidth]{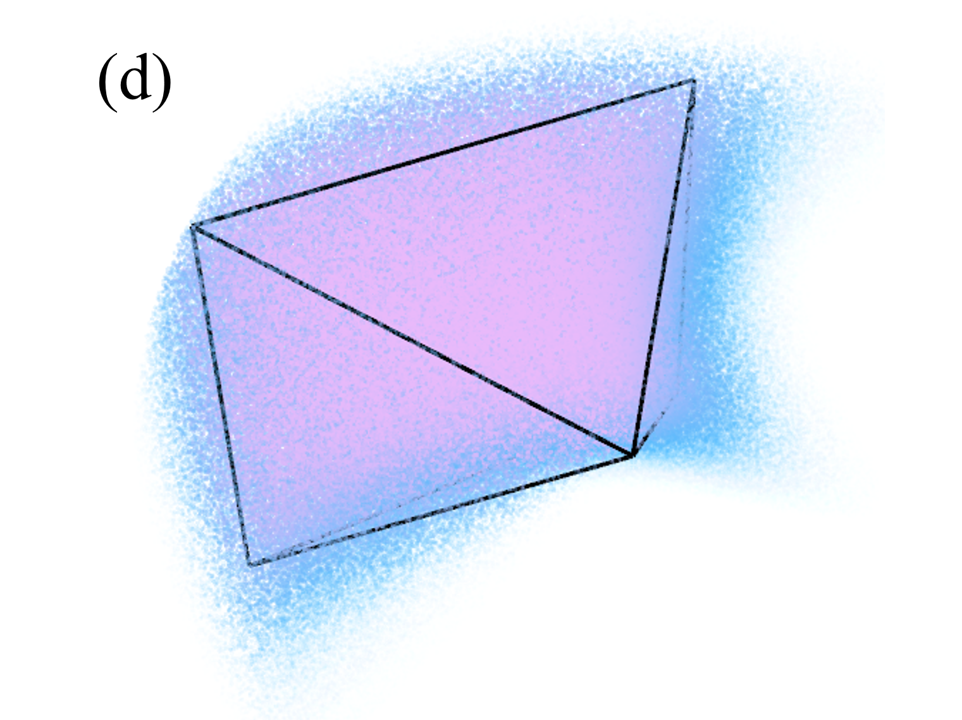} \label{fig:visib_model_4}}
\vspace{-0.8cm}
\caption{Illustration of the proposed visibility model under FoV setting $(\alpha_h, \alpha_v) = (90^\circ, 60^\circ)$. The violet shape in (a-c) represents visible regions. (d) visualizes the proposed visibility model by dense sampling. Points with visibility model take values close to zero are transparent and colored blue, and values close to one are opaque and colored pink.}
\vspace{-1.0cm}
\label{fig:visib_model}
\end{figure}

With the mutually orthogonal unit vectors $\vect{n}_1, \vect{n}_2, \vect{n}_3$ and the bearing vector $\vect{b} = \vect{f}_j - \vect{p}_i$, we can now define our differentiable visibility model as
\begin{align}
  v(\vect{s}_i, \vect{f}_j) &= v_1(\theta_1) \cdot v_3(\theta_3) \cdot v_2(\theta_2), \label{eq:visibility}\\
  v_1(\theta_1(i,j)) &= (1 + e^{-k_1(\sin\theta_1 - \sin\alpha_1)})^{-1}, \label{eq:v1}\\
  v_3(\theta_3(i,j)) &= (1 + e^{-k_3(\sin\theta_3 - \sin\alpha_3)})^{-1}, \label{eq:v3}\\
  v_2(\theta_2(i,j)) &= (1 + e^{-k_2(\cos\theta_2 - \cos\alpha_2)})^{-1}, \label{eq:v2}
\end{align}
where $\alpha_1 = (\pi - \alpha_v) \slash 2, \alpha_2 = \pi / 2, \alpha_3 = (\pi - \alpha_h) \slash 2$, and $\theta_1, \theta_2, \theta_3$ are the angles between the bearing vector $\vect{b}$ and the corresponding unit vectors $\vect{n}_1, \vect{n}_2, \vect{n}_3$ respectively as shown in Fig.\ref{fig:visib_model}(a-c). Then
\begin{equation} \label{eq:thetas}
  \sin\theta_1 = \frac{|\vect{n}_1 \times \vect{b}|}{|\vect{b}|}, \; \cos\theta_2 = \frac{\vect{n}_2 \cdot \vect{b}}{|\vect{b}|}, \; \sin\theta_3 = \frac{|\vect{n}_3 \times \vect{b}|}{|\vect{b}|}.
\end{equation}

Intuitively, the idea behind this visibility model construction is a deductive scheme as illustrated in Fig.~\ref{fig:visib_model}.
In the beginning, a sphere is used to represent omni-directional sensing within FoV depth.
Then the visibility components $v_1(\theta_1)$ and $v_3(\theta_3)$ carve out two pairs of axially symmetric cones in grey as illustrated in Fig.~\ref{fig:visib_model_1} and Fig.~\ref{fig:visib_model_2}, accounting for the vertical and horizontal FoV respectively.
Consequently, we name $v_1(\theta_1)$ and $v_3(\theta_3)$ as \textit{vertical visibility} and \textit{horizontal visibility}.
Lastly, $v_2(\theta_2)$ is used to choose the correct pyramidal shape among the remaining two according to $\theta_2$ resulting in Fig.~\ref{fig:visib_model_3}.
In (\ref{eq:v1})-(\ref{eq:v2}), we use sigmoid functions to approximate the step functions.
We illustrate (\ref{eq:v1}) and (\ref{eq:v2}) in Fig.~\ref{fig:sigmoid} for different $k$ and we opt for $(k_1, k_2, k_3)=(40, 10, 20)$ in our implementation.

Moreover, we can further induce the covisibility measure for states $\vect{s}_{1}$ and $\vect{s}_{2}$ as
\begin{equation} \label{eq:covisib}
  \mu(\vect{s}_{1}, \vect{s}_{2}) = \sum_{j\in J} v(\vect{s}_{1}, \vect{f}_j) v(\vect{s}_{2}, \vect{f}_j),
\end{equation}
where $J$ is the set of features within FoV depth of both states and $v(\vect{s}_{i}, \vect{f}_j)$ is the visibility model specified in (\ref{eq:visibility}).

We decompose the optimization of the covisibility measure into two phases based on the observation that $\theta_1$ is independent of the yaw direction.
The reason is that $\theta_1$ depends only on $\vect{n}_1$ and $\vect{b}$ according to (\ref{eq:thetas}), which can be completely determined by the position trajectory and its derivatives.
In the first phase where we optimize the position trajectory, we aim to maximize the \textit{vertical covisibility} considering $\theta_1$ to include as many covisible features as possible within the ring-shape FoV (Fig.\ref{fig:visib_model_1}) along the trajectory.
In the second phase, we optimize the yaw trajectory given the position trajectory, which yields the best sector (Fig.\ref{fig:visib_model_3}) to further maximize the covisibility measure $\mu$.
The two-phase strategy reduces the computational complexity of the covisibility optimization problem, facilitating efficient convergence towards high-quality trajectories.
Although this two-step optimization may not guarantee maximum covisibility, it is capable of generating a sufficiently informative trajectory to support accurate state estimation during maneuvers.

\subsection{Position Trajectory Optimization}
\label{subsec:pos_traj}
The position trajectory is represented using a degree $p=3$ uniform B-spline curve defined by $n+1$ control points $\{\vect{Q}_0, \vect{Q}_1, \cdots, \vect{Q}_n\}$ and a uniform knot span $\Delta t$, both of which serve as the variables in the position trajectory optimization.
The initial value of the trajectory is set as an initial B-spline curve that fits the collision-free path provided by standard method like A* and RRT*. The overall cost function objective is defined as
\begin{equation}
f_\text{total} = \lambda_\text{vc}f_\text{vc} + \lambda_\text{para} f_\text{para} + \lambda_\text{o}f_\text{o},
\end{equation}
where $f_\text{vc}$ and $f_\text{para}$ are the vertical covisibility and the parallax cost.
$f_\text{o}$ is the cost function for other factors including smoothness, safety, agility and dynamic feasibility.
These factors are traded off by $\lambda_\text{para}, \lambda_\text{vc}, \lambda_\text{o}$.

We define the vertical covisibility cost utilizing the angle $\theta_1$ in the visibility model introduced in Sec. \ref{subsec:model}.
For a pair of adjacent kont points $\vect{P}_i; \vect{P}_{i+1}$ (defined in Sec.\ref{subsec:bspline}) and a feature $\vect{f}_j$, we first calculate their vertical visibility angles $\theta_1(i, j)$ and $\theta_1(i+1, j)$ respectively.
Here, $\theta_1(k, j)$ is the angle between the thrust direction $\vect{n}_1(k)$ at knot points $\vect{P}_k$ and the bearing vector $\vect{b}(k, j) = \vect{f}_j - \vect{P}_k$, where $k = i, i+1$, i.e.
\begin{equation}
\theta_1(k, j) = \arccos\frac{\vect{n}_1(k)\cdot\vect{b}(k,j)}{|\vect{n}_1(k)||\vect{b}(k,j)|}.
\end{equation}
Then the vertical covisibility cost $f_\text{vc}$ is calculated as
\begin{equation}
f_\text{vc} = \sum_{i\in I}\sum_{j\in J_i} \underbrace{(1+g(\theta_1(i,j)))(1+g(\theta_1(i+1,j)))-1}_{h(i,j)},
\end{equation}
where $I$ is the set of knot points, $J_i$ is the set of features within FoV depth of both $\vect{P}_i$ and $\vect{P}_{i+1}$, and $g(\theta_1(i,j))$ is a differentiable potential function on $\theta_1(i,j)$ defined as
\begin{equation}
g(\theta_1(k, j)) = 
\begin{cases}
  a_1(\theta_1(k, j) - \theta_{1,\text{max}})^2, & \text{if } \theta_1(k, j) > \theta_{1,\text{max}} \\
  a_1(\theta_1(k, j) - \theta_{1,\text{min}})^2, & \text{if } \theta_1(k, j) < \theta_{1,\text{min}} \\
  0, & \text{otherwise}
\end{cases}
\end{equation}
where $\theta_{1,\text{max}} = (\pi - \alpha_v) \slash 2, \theta_{1,\text{min}} = (\pi + \alpha_v) \slash 2$, and $a_1$ is a scalar constant.
When the feature $\vect{f}_j$ is vertically covisible from both $\vect{P}_i$ and $\vect{P}_{i+1}$, the cost function $h(i,j)$ takes value zero.
Otherwise, it takes a positive value, penalizing the loss of vertical covisibility.

For the parallax cost $f_\text{para}$, we first calculate the parallax angle $\rho(i,j)$ for adjacent knot points $\vect{P}_i, \vect{P}_{i+1}$ and a feature $\vect{f}_j$ as
\begin{equation}
  \rho(i,j) = \arccos \frac{(\vect{P}_i-\vect{f}_j)\cdot(\vect{P}_{i+1}-\vect{f}_j)}{|\vect{P}_i-\vect{f}_j||\vect{P}_{i+1}-\vect{f}_j|}.
\end{equation}
Then the parallax cost $f_\text{para}$ is defined as
\begin{equation} \label{eq:f_para}
  f_\text{para} = \sum_{i\in I}\sum_{j\in J_i}w_{i,j}h(\rho(i,j)),
\end{equation}
where the constant weight $w_{i,j} = v_1^\prime(\theta_1(i,j))$ with parameter $k_1^\prime=60, \alpha_1^\prime = (\pi - 0.8\alpha_v)\slash 2$ (shown in Fig.~\ref{fig:sigmoid}) is used, giving lower weights to features near vertical FoV boundary and negligible weights to vertically invisible ones.
$h(\rho(i,j))$ is a differentiable potential function penalizing large parallax angle which exceeds threshold $\rho_\text{thr}$, 
\begin{equation}
h(\rho(i,j)) = \begin{cases}
  a_2 (\rho(i,j) - \rho_\text{thr})^2, & \text{if }\rho(i,j) > \rho_\text{thr} \\
  0, & \text{otherwise}
\end{cases}
\end{equation}
where $a_2$ is a scalar constant and $\rho_\text{thr} = \nu \rho_\text{max} \Delta t$ with image frame frequency $\nu$, the acceptable maximum parallax angle between consecutive frames $\rho_\text{max}$, and the knot span $\Delta t$.

The formulation of the cost function $f_\text{o}$ is similar to \cite{zhou2019robust} and \cite{zhou2020raptor}.
The smoothness, safety, agility and dynamic feasibility are achieved by minimizing jerk, collision risk, execution time and infeasible velocity and acceleration penalty respectively.

\subsection{Yaw Trajectory Generation}
\label{subsec:yaw_traj}
We first search a yaw primitives path (Sec.\ref{subsubsec:yaw_prim}), which serves as a guidance for the subsequent yaw trajectory optimization (Sec.\ref{subsubsec:yaw_opt}), avoiding local minima trap.

\subsubsection{Primitives Search}
\label{subsubsec:yaw_prim}
We model the problem as a graph search problem which seeks to maximize the cumulative covisibility measure along the trajectory while ensuring yaw dynamics feasibility.
At each knot point $\vect{P}_i \in \{\vect{P}_p, \vect{P}_{p+1}, \cdots, \vect{P}_{m-p}\}$ of the optimized position trajectory obtained from Sec.\ref{subsec:pos_traj}, a set of yaw angles $L_i = \{\psi_{i,0}, \psi_{i,1}, \cdots, \psi_{i,l}\}$ are sampled uniformly, each of which serves as a graph node.
An edge is built between two nodes $\psi_{i_1,j_1}$ and $\psi_{i_2,j_2}$ if and only if they belong to consecutive sets and fulfill the yaw dynamics, i.e.
$|i_1 - i_2| = 1$ and $|\psi_{i_1,j_1} - \psi_{i_2,j_2}| < \dot{\psi}_\text{max} \Delta t$.
For each edge, the gain associated with it is calculated as the covisibility measure according to (\ref{eq:covisib}).
Since there is no constraint on the start and end yaw values, we add two extra graph nodes $\psi_{p-1,0}, \psi_{m-p+1,0}$ and connect them with all nodes in set $L_p$ and $L_{m-p}$ respectively with zero-gain edges.
Then the problem is formulated as a shortest path problem from node $\psi_{p-1,0}$ to $\psi_{m-p+1,0}$, where the cost of each edge is the negative of the associated gain, and it can be efficiently solved by the Dijkstra algorithm.

\subsubsection{Trajectory Optimization}
\label{subsubsec:yaw_opt}
The yaw trajectory $\psi(t)$ is also represented as a degree $p=3$ uniform B-spline curve defined by a same number of control points as the position trajectory $\Xi = \{\xi_0, \xi_1, \cdots, \xi_n\}$ and a same knot span $\Delta t$.
Then the yaw knot points $\{\psi_p, \psi_{p+1}, \cdots, \psi_{m-p}\}$ correspond to the same time instance to the yaw primitives $\Psi^* = \{\psi^*_p, \psi^*_{p+1}, \cdots, \psi^*_{m-p}\}$ obtained from Sec.\ref{subsubsec:yaw_prim}.
Then the optimization problem is formulated as
\begin{equation}
  \argmin_{\Xi} \int_{t_p}^{t_{m-p}} \dddot{\psi}(t)^2 dt + \sum_{i\in I} (\lambda_1 (\psi_i-\psi_i^*)^2 - \lambda_2  \mu(\vect{s}_i, \vect{s}_{i+1})),
\end{equation}
Here the first term minimizes the jerk and ensures smoothness.
The second term accounts for yaw primitives guidance.
For the covisibility term, the definitions (\ref{eq:visibility})(\ref{eq:covisib}) gives
\begin{equation}
  \begin{aligned}
    \mu(\vect{s}_i, \vect{s}_{i+1}) &= \sum_{j\in J_i} v(\vect{s}_{i}, \vect{f}_j) v(\vect{s}_{i+1}, \vect{f}_j) \\
    &= \sum_{j\in J_i} \underbrace{v_1(i) v_1(i+1) v_2(i) v_2(i+1)}_\text{constant} v_3(i)  v_3(i+1),
  \end{aligned}
\end{equation}
where we write $v_k(\theta_k(i,j))$ as $v_k(i)$ for simplicity without ambiguity.
Thanks to the decomposable visibility model, we can treat the first four terms of the covisibility as a constant and only optimize the last two terms with respect to yaw.

\begin{table}[t]
  \caption{Estimation Accuracy Benchmark}
      \begin{tabular}{cccccm{0.3cm}m{0.3cm}m{0.3cm}m{0.3cm}}
     \hline\hline
     \textbf{Top} & \multirow{2}{*}{\textbf{Method}} & \textbf{Path}  & \textbf{Time} & \textbf{Fail}      & \multicolumn{2}{c}{\textbf{GE(m)$^\dagger$}} & \multicolumn{2}{c}{\textbf{RMSE(m)}} \\
     \textbf{Vel} &                                  & \textbf{(m)}   & \textbf{(s)$^\diamond$}  & \textbf{Rate$^\ddagger$}      & \textbf{Avg}       & \textbf{Std}    & \textbf{Avg}       & \textbf{Std}          \\
     \hline
     \multirow{3}{*}{\begin{tabular}[c]{@{}c@{}}$2.0$\\$m\slash s$\end{tabular}}            & PAG\cite{zhou2019robust}         & 18.3  & \textbf{0.05} &  10/10         & - & -             & - & - \\
                                       & PAW\cite{Murali2019}             & 19.9  & 4.17  & 1/10          & 3.23 & 2.36      & 1.69 & 1.16            \\
                                       & Ours                             & 19.9  & 0.24  & \textbf{0/10}          & \textbf{0.12} & \textbf{0.04}    & \textbf{0.13} & \textbf{0.03} \\
     \hline
     \multirow{3}{*}{\begin{tabular}[c]{@{}c@{}}$4.0$\\$m\slash s$\end{tabular}}            & PAG\cite{zhou2019robust}         & 18.6  & \textbf{0.11} & 10/10          & - & -             & - & - \\
                                       & PAW\cite{Murali2019}             & 19.9  & 4.17 & 8/10          & 7.98 & 1.75      & 4.15 & 1.06             \\
                                       & Ours                             & 20.9  & 0.25 & \textbf{0/10} & \textbf{0.14} & \textbf{0.09}       & \textbf{0.16} & \textbf{0.04}              \\
     \hline\hline
     \vspace{-0.2cm}
    \end{tabular}
    \footnotesize{$^\diamond$ Average trajectory generation time.}\\
    \footnotesize{$^\dagger$ GE abbreviates for goal error.}\\
    \footnotesize{$^\ddagger$ If GE exceeds 10m, the task is deemed fail and the result will not be counted towards GE and RMSE statistics.}\\

  \vspace{-2.0cm}
  \label{tab:result}
\end{table}

\section{Experiments}
\label{sec:experiments}

\subsection{Experimental Setup}
For simulation experiments, we utilize a photorealistic simulator AirSim \cite{shah2018airsim, han2021fast}.
All simulation experiments have been run on a computer equipped with an Intel Core i7-13700F, GeForce RTX 3060 12G and 32GB memory with identical input feature cloud.
For real-world experiments, we use a custom built light-weight and high-agility quadrotor platform, which has a take-off weight of $861$g and thrust-to-weight ratio $\sim 5.3$.
The quadrotor is equipped with an NVIDIA Jetson Orin NX 16 GB and an NxtPx4 autopilot\footnote{\url{https://github.com/HKUST-Aerial-Robotics/NxtPX4}}.
The visual data is provided by a forward-looking global-shutter camera RealSense\footnote{\url{https://www.intelrealsense.com/depth-camera-d435}} with $86^\circ$ horizontal FoV and $57^\circ$ vertical FoV.
The odometry ground truth is provided by OptiTrack motion capture system, which is also utilized as controller feedback.
In both simulation and real-world experiments, we adopt VINS-Fusion \cite{qin2019general} as our visual state estimator, where inertial measurement unit (IMU) information is excluded to eliminate its contribution to the visual odometry (VO).
For implementation details, we use $a_1 = 10, a_2 = 20, \nu = 20\text{Hz}, \rho_\text{max} = 10^\circ$ and solve the optimization with a general nonlinear solver\footnote{\url{https://nlopt.readthedocs.io/en/latest}}.


\begin{figure}[t]
	\centering
  \includegraphics[width=\columnwidth]{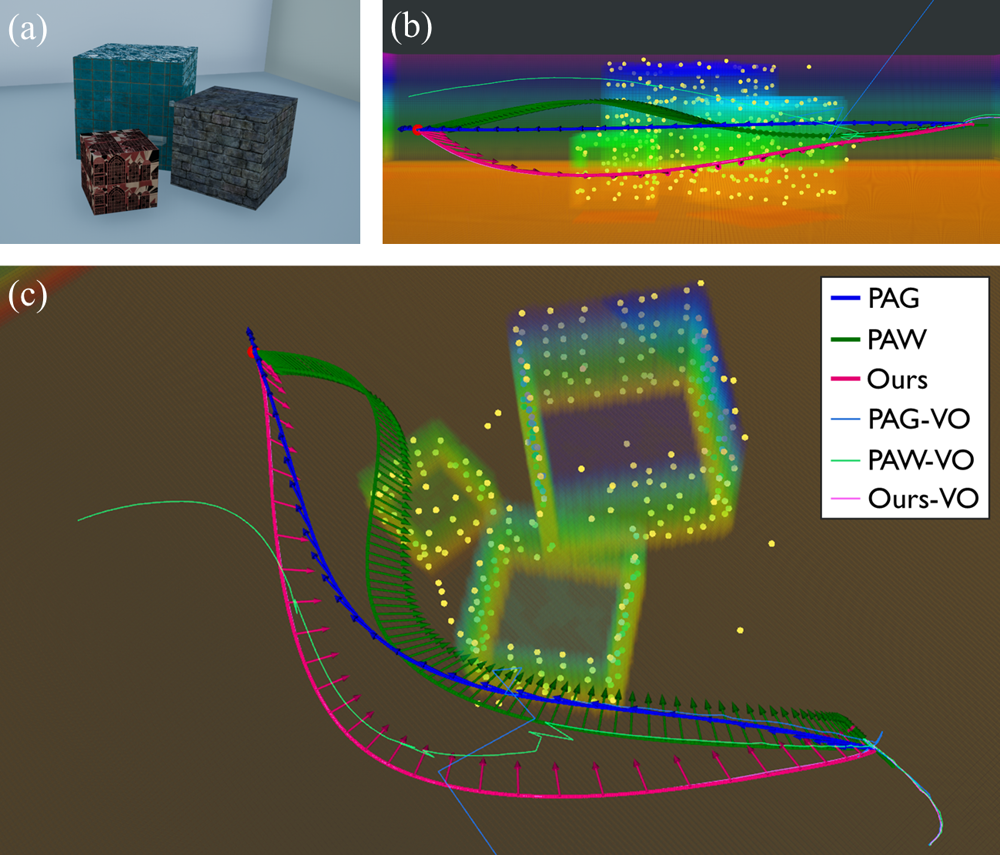}  
  \caption{The setup in the photorealistic simulator Airsim is shown in (a). The trajectories generated by PAG\cite{zhou2019robust}, PAW\cite{Murali2019} with top velocity $2.0 m\slash s$ and our planner with top velocity $4.0 m\slash s$ along with the visual odometries (VO) are illustrated from side and top views in (b) and (c) respectively. The yaw directions are visualized as arrows at interval of 0.2s.}
  \label{fig:simulation1}
  \vspace{-0.2cm}
\end{figure}

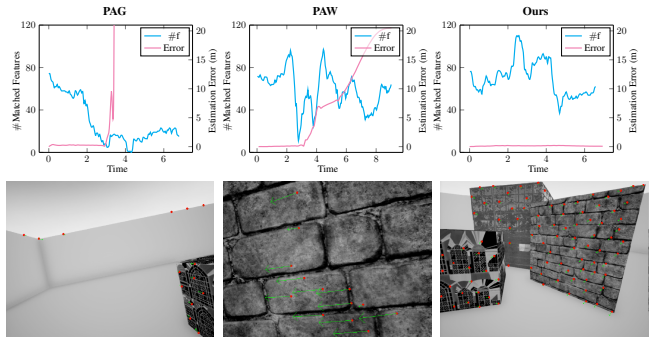
\begin{figure}[t]
  \centering
	\subfigtopskip=2pt
	\subfigbottomskip=2pt
	\subfigcapskip=-5pt
  \resizebox{0.33\columnwidth}{!}{%
    \begin{tikzpicture}
      \begin{axis}[
        scale=0.8, 
        legend style={fill=white, fill opacity=0.6, draw opacity=1, text opacity=1},
        ymin=0, ymax=120, xmin=-0.5, xmax=7.5, 
        axis y line* = left, 
        ytick={0,40,...,120}]
        \addplot[color=Cyan,smooth,very thick] table [x=x, y=y, col sep=comma] {feature_cnt_plot_pag_min5.csv}; \label{plot_1}
      \end{axis}
      \begin{axis}[
        scale=0.8, 
        legend style={fill=white, fill opacity=0.6, draw opacity=1, text opacity=1},
        ymin=-1, ymax=21, xmin=-0.5, xmax=7.5,
        hide x axis,
        axis y line*=right]
        \addlegendimage{/pgfplots/refstyle=plot_1}\addlegendentry{$\#$f}
        \addplot[color=CarnationPink,smooth,very thick] table [x=x, y=y, col sep=comma] {error_plot_pag.csv};
        \addlegendentry{Error}
      \end{axis}
      \node at (2.8cm,-0.6cm) {Time};
      \node at (-0.7cm,2.2cm) [rotate=90, anchor=base] {$\#$Matched Features};
      \node at (6.3cm,2.2cm) [rotate=90, anchor=base] {Estimation Error (m)};
      \node[above,font=\large\bfseries] at (current bounding box.north) {PAG};
    \end{tikzpicture}
  } \hskip -5pt
  \resizebox{0.33\columnwidth}{!}{%
    \begin{tikzpicture}
      \begin{axis}[
        scale=0.8, 
        legend style={fill=white, fill opacity=0.6, draw opacity=1, text opacity=1},
        ymin=0, ymax=120, xmin=-0.5, xmax=9.8, 
        axis y line* = left, 
        ytick={0,40,...,120}]
        \addplot[color=Cyan,smooth,very thick] table [x=x, y=y, col sep=comma] {feature_cnt_plot_bmk_min5.csv}; \label{plot_1}
      \end{axis}
      \begin{axis}[
        scale=0.8, 
        legend style={fill=white, fill opacity=0.6, draw opacity=1, text opacity=1},
        ymin=-1, ymax=21, xmin=-0.5, xmax=9.8,
        hide x axis,
        axis y line*=right]
        \addlegendimage{/pgfplots/refstyle=plot_1}\addlegendentry{$\#$f}
        \addplot[color=CarnationPink,smooth,very thick] table [x=x, y=y, col sep=comma] {error_plot_bmk.csv};
        \addlegendentry{Error}
      \end{axis}
      \node at (2.8cm,-0.6cm) {Time};
      \node at (-0.7cm,2.2cm) [rotate=90, anchor=base] {$\#$Matched Features};
      \node at (6.3cm,2.2cm) [rotate=90, anchor=base] {Estimation Error (m)};
      \node[above,font=\large\bfseries] at (current bounding box.north) {PAW};
    \end{tikzpicture}
  } \hskip -5pt
  \resizebox{0.33\columnwidth}{!}{%
    \begin{tikzpicture}
      \begin{axis}[
        scale=0.8, 
        legend style={fill=white, fill opacity=0.6, draw opacity=1, text opacity=1},
        ymin=0, ymax=120, xmin=-0.5, xmax=7.5, 
        axis y line* = left, 
        ytick={0,40,...,120}]
        \addplot[color=Cyan,smooth,very thick] table [x=x, y=y, col sep=comma] {feature_cnt_plot_ours_min5.csv}; \label{plot_1}
      \end{axis}
      \begin{axis}[
        scale=0.8, 
        legend style={fill=white, fill opacity=0.6, draw opacity=1, text opacity=1},
        ymin=-1, ymax=21, xmin=-0.5, xmax=7.5,
        hide x axis,
        axis y line*=right]
        \addlegendimage{/pgfplots/refstyle=plot_1}\addlegendentry{$\#$f}
        \addplot[color=CarnationPink,smooth,very thick] table [x=x, y=y, col sep=comma] {error_plot_ours.csv};
        \addlegendentry{Error}
      \end{axis}
      \node at (2.8cm,-0.6cm) {Time};
      \node at (-0.7cm,2.2cm) [rotate=90, anchor=base] {$\#$Matched Features};
      \node at (6.3cm,2.2cm) [rotate=90, anchor=base] {Estimation Error (m)};
      \node[above,font=\large\bfseries] at (current bounding box.north) {Ours};
    \end{tikzpicture}
  } \\
  \subfigure{\includegraphics[width=0.32\columnwidth]{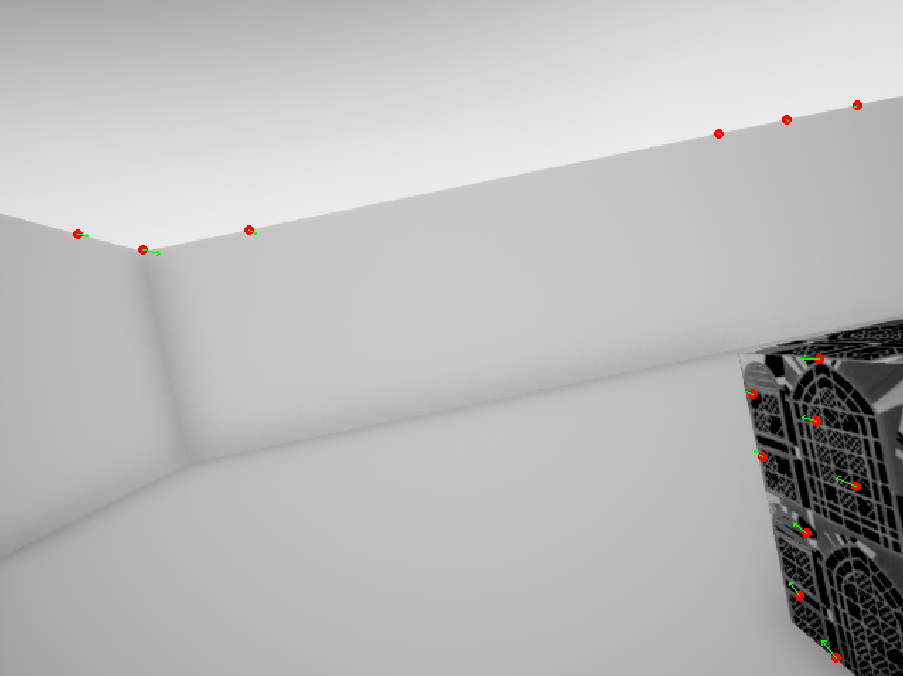}}
  \subfigure{\includegraphics[width=0.32\columnwidth]{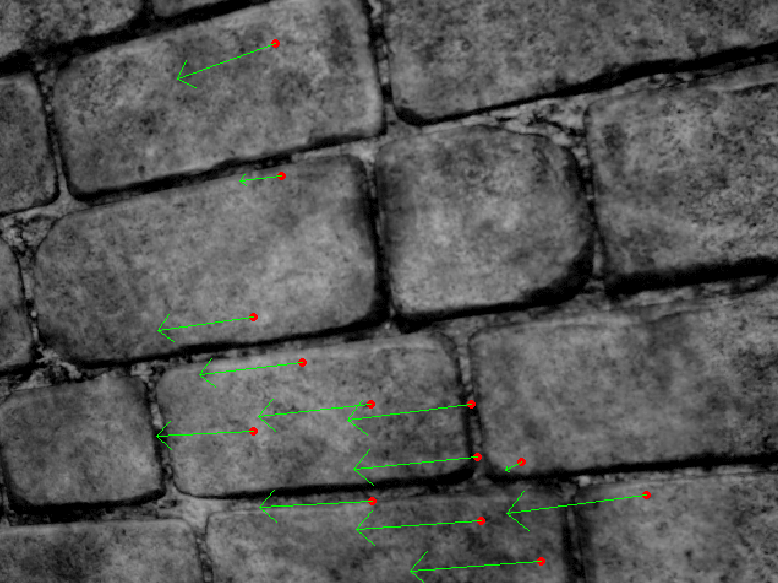}}
  \subfigure{\includegraphics[width=0.32\columnwidth]{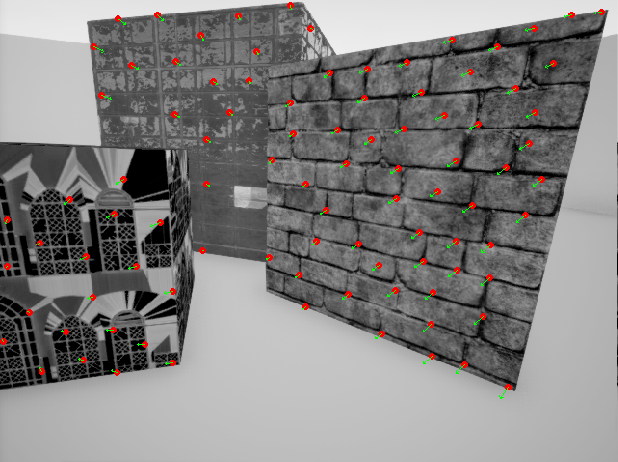}} \\
\caption{The top row shows plots of matched features number and state estimation error over time for the three trajectories, i.e. PAG\cite{zhou2019robust}, PAW\cite{Murali2019} and ours, at a top velocity of $4.0 m\slash s$ repectively from left to right. The bottom row showcases corresponding snapshots of quadrotor's first-person view, where the red dots represent the matched features and the green arrows indicate the pixel of matched features from the previous frame.}
  \vspace{-0.6cm}
  \label{fig:simulation2}
\end{figure}

\subsection{Photorealistic Simulation Benchmarks}
We perform benchmark comparison with a perception-agnostic (PAG) planner \cite{zhou2019robust} and a state-of-the-art perception-aware (PAW) planner \cite{Murali2019}.
As there is no open-source code available for \cite{Murali2019}, we use our implementation.
The experiment setup consists of a pile of texture-rich boxes in the middle of a textureless room as shown in Fig.\ref{fig:simulation1}.
The quadrotor is commanded to execute each trajectory in different agilities in terms of top velocity for $10$ runs and the result statistics are shown in Table~\ref{tab:result}.
We also illustrate the generated trajectories along with those estimated by VO in Fig.\ref{fig:simulation1}.
As the position trajectories from \cite{zhou2019robust} \cite{Murali2019} are not optimized for perceptual objectives, they tend to take short routes which pass too close to the grid-texture grey box resulting in large parallax angles.
The perception-aware yaw trajectory from \cite{Murali2019} barely allowed task completion, yielding poor localization accuracy, while the head-forwarding yaw trajectory from \cite{zhou2019robust} caused the quadrotor to face the white wall, leading to task failure.
In contrast, the trajectory given by us is able to maintain a reasonable distance from the grey box to ensure small enough parallax angles and includes the small red box in the FoV to ensure sufficient covisible features for localization. 
This resulted in a sufficient and stable number of matched features as shown in Fig.~\ref{fig:simulation2} and consequently accurate state estimation with low goal error and RMSE.
Moreover, our method is capable of generating trajectories efficiently in less than $0.25$s, thanks to the decomposable visibility model, while \cite{Murali2019} suffers from computational complexity due to a large number of features.

\subsection{Real-world Experiments}
\begin{figure}[t]
  \centering
	\subfigtopskip=2pt
	\subfigbottomskip=2pt
	\subfigcapskip=-3pt
  \subfigure{\includegraphics[width=0.85\columnwidth]{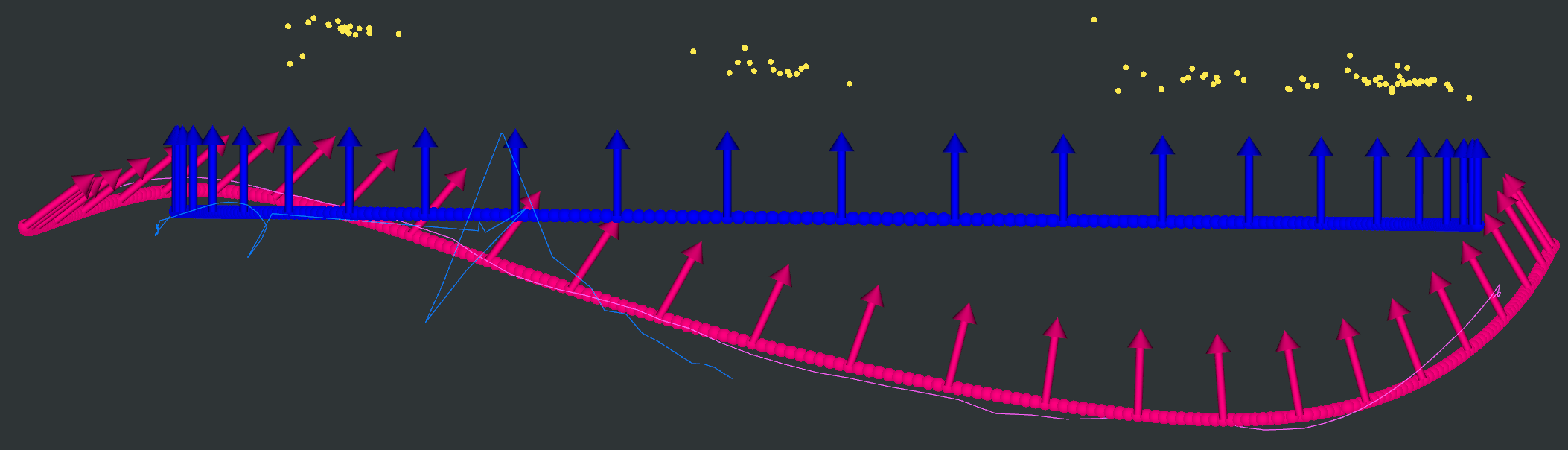}}
  \subfigure{\includegraphics[width=0.85\columnwidth]{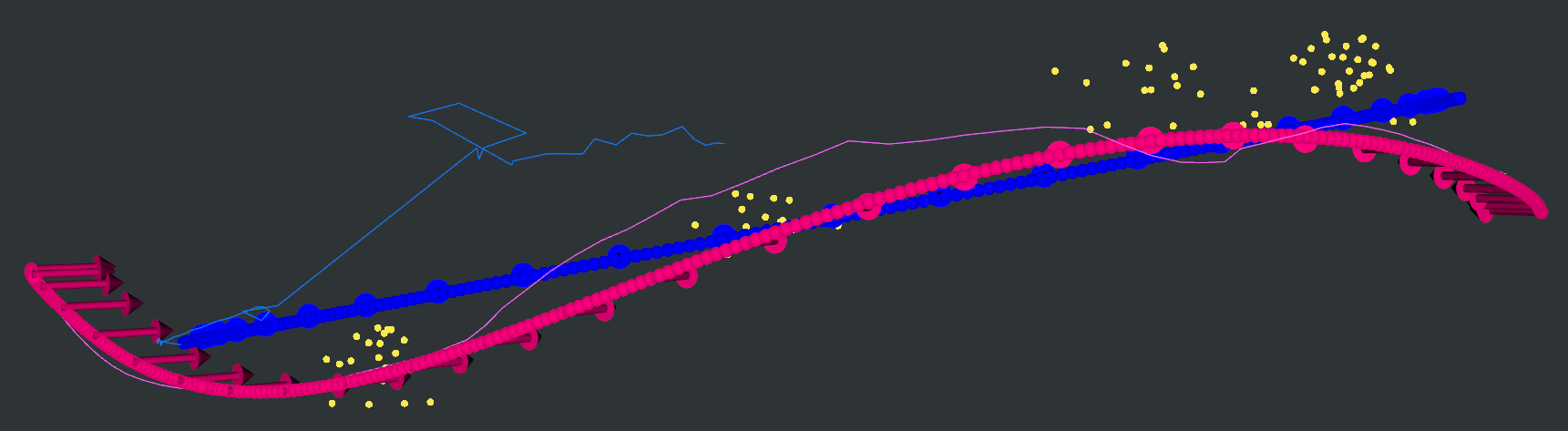}}
  \resizebox{0.49\columnwidth}{!}{%
    \begin{tikzpicture}
      \begin{axis}[
        scale=0.9, 
        legend style={fill=white, fill opacity=0.6, draw opacity=1, text opacity=1},
        ymin=0, ymax=50, xmin=-0.5, xmax=10.5, 
        axis y line* = left, 
        ytick={0,10,...,50}]
        \addplot[color=Cyan,smooth,very thick] table [x=x, y=y, col sep=comma] {real_feature_cnt_plot_ours.csv}; \label{plot_1}
      \end{axis}
      \begin{axis}[
        scale=0.9, 
        legend style={fill=white, fill opacity=0.6, draw opacity=1, text opacity=1},
        ymin=-0, ymax=4.5, xmin=-0.5, xmax=10.5,
        hide x axis,
        axis y line*=right]
        \addlegendimage{/pgfplots/refstyle=plot_1}\addlegendentry{$\#$f}
        \addplot[color=CarnationPink,smooth,very thick] table [x=x, y=y, col sep=comma] {real_error_plot_ours.csv};
        \addlegendentry{Error}
      \end{axis}
      \node at (3cm,-0.7cm) {Time};
      \node at (-0.7cm,2.5cm) [rotate=90, anchor=base] {$\#$Matched Features};
      \node at (7.0cm,2.5cm) [rotate=90, anchor=base] {Estimation Error (m)};
      \node[above,font=\large\bfseries] at (current bounding box.north) {Ours};
    \end{tikzpicture}
  } \hskip -5pt
  \resizebox{0.49\columnwidth}{!}{%
    \begin{tikzpicture}
      \begin{axis}[
        scale=0.9, 
        legend style={fill=white, fill opacity=0.6, draw opacity=1, text opacity=1},
        ymin=0, ymax=50, xmin=-0.5, xmax=8.0, 
        axis y line* = left, 
        ytick={0,10,...,50}]
        \addplot[color=Cyan,smooth,very thick] table [x=x, y=y, col sep=comma] {real_feature_cnt_plot_line.csv}; \label{plot_1}
      \end{axis}
      \begin{axis}[
        scale=0.9, 
        legend style={fill=white, fill opacity=0.6, draw opacity=1, text opacity=1},
        ymin=-0, ymax=4.5, xmin=-0.5, xmax=8.0,
        hide x axis,
        axis y line*=right]
        \addlegendimage{/pgfplots/refstyle=plot_1}\addlegendentry{$\#$f}
        \addplot[color=CarnationPink,smooth,very thick] table [x=x, y=y, col sep=comma] {real_error_plot_line.csv};
        \addlegendentry{Error}
      \end{axis}
      \node at (2.8cm,-0.6cm) {Time};
      \node at (-0.7cm,2.5cm) [rotate=90, anchor=base] {$\#$Matched Features};
      \node at (7.0cm,2.5cm) [rotate=90, anchor=base] {Estimation Error (m)};
      \node[above,font=\large\bfseries] at (current bounding box.north) {Perception-Agnostic};
    \end{tikzpicture}
  }
  \vspace{-0.4cm}
\caption{The figures show the trajectory generated by our methods in pink and a perception-agnostic trajectory in blue at top velocity of $3.7m\slash s$ along with their VO results from top and side views.
The start and goal of the perception-agnostic trajectory have been adjusted to guarantee existence of visible features at these two points.
The bottom row shows plots of matched features number and state estimation error over time for the two trajectories.}
\vspace{-0.8cm}
\label{fig:real_world_combo}
\end{figure}

We further conduct experiments with a real quadrotor in a challenging darkness scenario with four illuminated spots distributed at different height levels as depicted in Fig.\ref{fig:realworld_large}.
Without perception-aware planning, a visual state estimator would not be able to sustain reliable performance in such a severe environment since the VO would start drifting drastically once the quadrotor encounters blindly complete darkness without any illuminated features.
As exemplified in Fig.~\ref{fig:real_world_combo}, we showcase a perception-agnostic trajectory and the plots reveal that it encounters zero features, resulting in task failure.
In contrast, our perception-aware trajectory of $11.3$m length maintains a sufficient number of matched features and yields accurate VO with an average goal error of $0.54$m and RMSE of $0.40$m in $5$ runs.
The results validate the feasibility of our perception-aware planner under challenging low-texture environments, thereby expanding the application boundaries of visual state estimation techniques.


\section{Conclusions}
This paper presents a perception-aware trajectory generation method for quadrotors aggressive flight.
Our approach takes the feature matchability into consideration by maximizing the number of covisible features and maintaining small-enough parallax angles.
Leveraging the decomposable visibility model proposed, we adopt a two-stage strategy to generate perception-aware trajectories that enhances visual-based estimator accuracy while satisfying the constraints on smoothness, safety, agility and the quadrotor dynamics.
Our method is validated in photorealistic simulations and real-world experiments, demonstrating significant improvement in state estimation accuracy with root mean square error (RMSE) reduced by up to an order of magnitude.
\addtolength{\textheight}{0.cm}   

\newlength{\bibitemsep}\setlength{\bibitemsep}{0.0\baselineskip}
\newlength{\bibparskip}\setlength{\bibparskip}{0.1pt}
\let\oldthebibliography\thebibliography
\renewcommand\thebibliography[1]{%
\oldthebibliography{#1}%
\setlength{\parskip}{\bibitemsep}%
\setlength{\itemsep}{\bibparskip}%
}

\bibliography{ref}

\end{document}